%% file: main.tex
\documentclass[10pt,twocolumn,letterpaper]{article}

\usepackage[pagenumbers]{cvpr} %

\input{preamble}
\input{macros}

\definecolor{cvprblue}{rgb}{0.21,0.49,0.74}
\usepackage[pagebackref,breaklinks,colorlinks,citecolor=cvprblue]{hyperref}
\usepackage{cuted}

\title{\paperTitle}

\author{\authorBlock}
\begin{document}
\maketitle
\begin{strip}\centering
\input{figures/01_teaser/figure}
\end{strip}
\input{sections/00_abstract}    
\input{sections/01_intro}

\input{sections/02_related}
\input{sections/04_method}
\input{sections/05_experiments}

\input{sections/06_conclusion}
\input{sections/08_ack}
{
    \small
    \setlength{\bibsep}{0pt}
    \bibliographystyle{abbrvnat}
    \bibliography{shortstrings, references}
}
\end{document}

%% file: preamble.tex
\usepackage[dvipsnames]{xcolor}
\usepackage{times}
\usepackage{microtype}
\usepackage{epsfig}
\usepackage{xcolor}
\usepackage{caption}
\usepackage{float}
\usepackage{placeins}
\usepackage{color, colortbl}
\usepackage{stfloats}
\usepackage{enumitem}
\usepackage{tabularx}
\usepackage{xstring}
\usepackage{multirow}
\usepackage{xspace}
\usepackage{cuted} %
\usepackage{url}
\usepackage{subcaption}
\usepackage{xcolor}
\usepackage[hang,flushmargin]{footmisc}
\usepackage{bm}
\usepackage{amsmath}
\usepackage{booktabs} %
\usepackage{makecell} %
\usepackage{amssymb}
\usepackage[normalem]{ulem}
\usepackage{makecell}

%% file: macros.tex
\def\paperID{}
\def\confName{CVPR}
\def\confYear{2026}

\def\modelname{Vanast\xspace}
\def\paperTitle{\textit{\modelname}: Virtual Try-On with Human Image Animation\\via Synthetic Triplet Supervision}

\def\authorBlock{
    Hyunsoo Cha \qquad
    Wonjung Woo \qquad
    Byungjun Kim \qquad
    Hanbyul Joo \\
    Seoul National University\\
    {\tt\small \{243stephen, agnes2327, byungjun.kim, hbjoo\}@snu.ac.kr} \\
    {\tt\small \href{https://hyunsoocha.github.io/vanast/}{\color{magenta}{https://hyunsoocha.github.io/vanast/}}}
    \vspace{-5mm}
}

\newif\ifreview 
\newif\ifarxiv 
\newif\ifcamera 
\newif\ifrebuttal 
\ifcamera \usepackage[accsupp]{axessibility} \fi
\ifcamera  \fi
\usepackage{subcaption} %
\ifarxiv  \fi

\definecolor{darkgreen}{RGB}{0,100,0}

\definecolor{tabfirst}{rgb}{1, 0.7, 0.7} %
\definecolor{tabsecond}{rgb}{1, 0.85, 0.7} %
\definecolor{tabthird}{rgb}{1, 1, 0.7} %

\newcommand{\ourmethod}[1]{our method}

\newcommand{\mat}[1]{\mathbf{#1}}

\newcommand{\R}[1]{{%
    \textbf{%
        \ifstrequal{#1}{1}{\textcolor{orange}{R#1}}{%
        \ifstrequal{#1}{2}{\textcolor{blue}{R#1}}{%
        \ifstrequal{#1}{3}{\textcolor{magenta}{R#1}}{%
        \ifstrequal{#1}{4}{\textcolor{teal}{R#1}}{%
                           \textcolor{cyan}{R#1}%
        }}}}%
    }%
}}

\definecolor{deepblue}{RGB}{83, 124, 186}

%% file: figures/01_teaser/figure.tex
\includegraphics[width=\linewidth, trim={0 0cm 0 0.0cm}, clip]{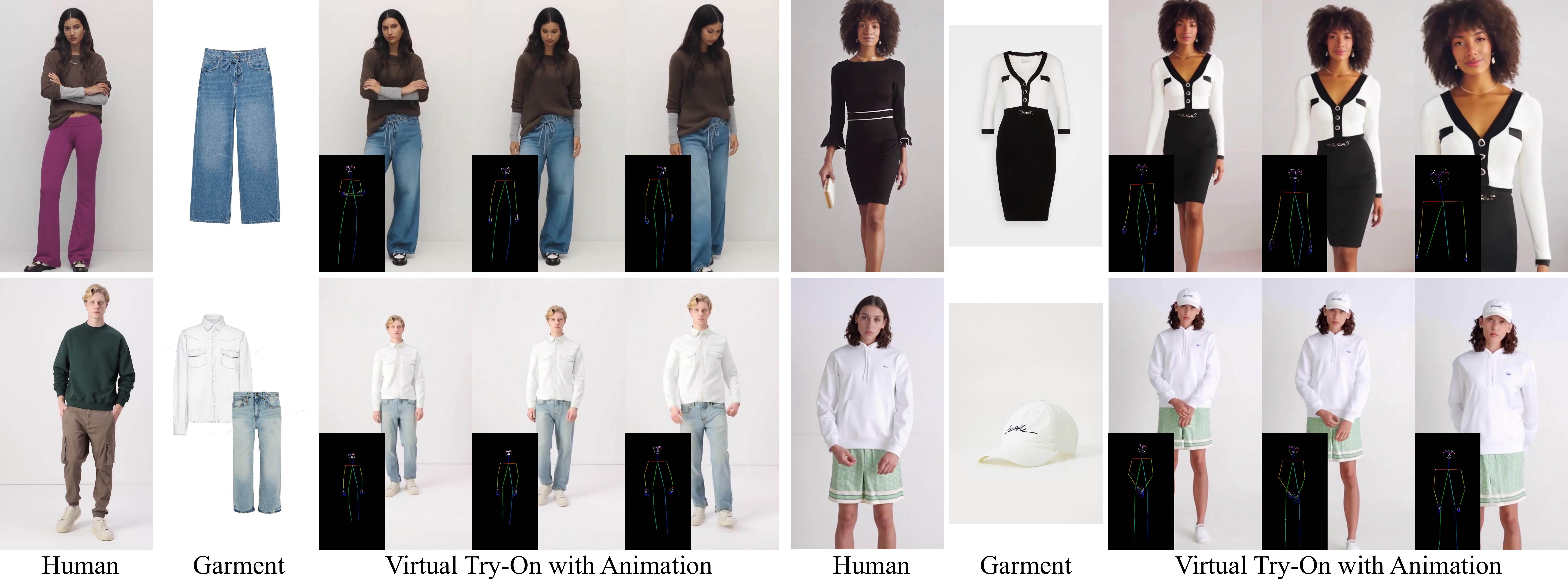}
\vspace{-13px}
\captionof{figure}{\textbf{\modelname.} 
Given a human image and one or more garment images, our method generates virtual try-on with human image animation conditioned on a pose video while preserving identity.
}
\vspace{-5px}
\label{fig:teaser}

%% file: sections/00_abstract.tex
\begin{abstract}
We present \modelname, a unified framework that generates garment-transferred human animation videos directly from a single human image, garment images, and a pose guidance video. 
Conventional two-stage pipelines treat image-based virtual try-on and pose-driven animation as separate processes, which often results in identity drift, garment distortion, and front–back inconsistency. Our model addresses these issues by performing the entire process in a single unified step to achieve coherent synthesis. 
To enable this setting, we construct large-scale triplet supervision.  Our data generation pipeline includes generating identity-preserving human images in alternative outfits that differ from garment catalog images, capturing full upper and lower garment triplets to overcome the single-garment–posed video pair limitation, and assembling diverse in-the-wild triplets without requiring garment catalog images.
We further introduce a Dual Module architecture for video diffusion transformers to stabilize training, preserve pretrained generative quality, and improve garment accuracy, pose adherence, and identity preservation while supporting zero-shot garment interpolation. Together, these contributions allow \modelname to produce high-fidelity, identity-consistent animation across a wide range of garment types.
\end{abstract}

%% file: sections/01_intro.tex
\vspace{-10px}
\section{Introduction}
A fundamental question arises when aiming to generate a garment-transferred human animation video: \textit{how can one synthesize a realistic video of a person wearing a target garment given only a single human image, one or more garment images, and a pose guidance video?}
A straightforward approach to building such a system with existing methods is to first apply an image-based virtual try-on model to a human image and garment images~\cite{kim2024stableviton, chong2024catvton, guo2025any2anytryon, feng2025omnitry}, then animate the synthesized result using a pose-driven video generation model~\cite{tu2025stableanimator, hu2024animate, zhu2024champ, li2024dispose}. While such a two-stage pipeline can produce plausible videos, it suffers from several inherent limitations. First, discrepancies between the training distributions of image try-on and video animation models cause identity drift, garment distortion, and compounding artifacts at inference time. Second, the decomposition into two separate models is computationally inefficient. Third, garments have distinct front–back geometry, yet standard video animation models operate from a single static image and therefore lose information required to synthesize consistent appearances across diverse viewpoints. These issues indicate that high-quality virtual try-on animation requires a unified, single-stage generation framework. Motivated by these challenges, we introduce \modelname, an end-to-end system that directly synthesizes garment-transferred human animation videos from a human image, one or more garment images, and a pose guidance video.

Building such a model requires triplet supervision consisting of a human image, garment images, and a RGB video of the same person moving while wearing the target garment. However, no existing public dataset provides this structure. Garment images and fashion videos can be collected from online retail sources, but the human image cannot simply be a frame sampled from the video. If the human image shows the same clothing as the video, the model learns to prioritize motion reenactment rather than garment transfer. The human image must therefore depict the person wearing different clothing from the video garment, a configuration that current datasets do not offer~\cite{dong2019vvt, fang2024vivid, nguyen2025swifttry}. To address this gap, we present a method to generate human images in alternative outfits while preserving identity. Our method transforms a given human image into a photorealistic version wearing garments different from those in the video, enabling the construction of accurate triplet pairs. To overcome the limitation of online shopping videos, where each clip typically features only one garment category, we additionally capture a dataset containing full upper and lower garment triplets.

Yet two challenges remain. 
First, transferring garments from real in-the-wild images is difficult because online catalog-style garments differ greatly from casually worn garments in unconstrained environments. Second, captured triplets suffer from limited scene diversity, leading to degradation when models are deployed outside their narrow capture conditions. To overcome these limitations, we introduce a scalable pipeline for constructing triplets from in-the-wild videos that lack garment images. This pipeline discovers garments, filters motion segments, and enforces identity consistency without requiring paired catalog images, expanding both visual diversity and garment variability.

A straightforward approach to train on such triplets is to modify a video diffusion transformer~\cite{wan2025wan} by feeding the human image, garment images, and pose video as additional conditioning signals. One option is to concatenate or fuse these inputs at the token level, but this severely hinders fast convergence and destabilizes optimization~\cite{jiang2025vace}. An alternative is to introduce a dedicated context module and fine-tune it to encode all conditions, yet this approach often fails to propagate the full set of constraints uniformly into the generated video, causing certain conditions to be underrepresented or ignored. To address these challenges, we propose a \textbf{Dual Module} architecture that preserves the pretrained text-to-video backbone while introducing dedicated pathways for garment transfer and pose guidance while preserving identity. This design converges quickly, maintains the generative fidelity of the original model, and significantly improves garment accuracy, pose adherence, and identity preservation. It further enables zero-shot garment interpolation through its modular conditioning structure, allowing smooth transitions between garment styles without additional finetuning while preserving identity.

In summary, our contributions are as follows:
\begin{itemize}
    \item We introduce the first unified framework, \modelname, that directly synthesizes human image animation videos with virtual try-on from a single human image, garment images, and a pose guidance video, eliminating the limitations of two-stage virtual try-on pipelines.
    \item We construct scalable triplet supervision for training by (1) generating identity-preserving human images wearing alternative garments, (2) capturing full upper and lower garment triplets to overcome single-garment constraints of online sources, and (3) building diverse in-the-wild triplets without requiring garment images.
    \item We propose a Dual Module architecture for video DiT that enables fast and stable convergence, preserves pretrained generative quality, and enhances garment transfer accuracy, pose adherence, and identity preservation, while also supporting zero-shot garment interpolation.
\end{itemize}

%% file: sections/02_related.tex
\section{Related Work}
\noindent \textbf{Virtual Try-On.}
Virtual try-on has evolved from geometry-driven clothing warping to correspondence-based synthesis that models fine-grained human–garment interactions. Early approaches rely on human and clothing parsing combined with geometric alignment and appearance blending~\cite{kim2024stableviton, Cha_2024_CVPR, yang2020acgpn}. While effective under well-aligned conditions, their dependence on explicit 2D warping often limits robustness in the presence of large pose variation, occlusions, or non-rigid garment deformation.
Recent diffusion-based methods have reshaped the landscape of image-based virtual try-on by replacing hand-designed warping modules with learned correspondence priors. Mask-conditioned dual-UNet architectures~\cite{xu2025ootdiffusion, chong2024catvton, Cha_2025_CVPR, cha2025durian} employ segmentation masks to enforce explicit spatial control and enhance compositional editing. Meanwhile, transformer-based diffusion models~\cite{feng2025omnitry, guo2025any2anytryon} leverage global self-attention to implicitly infer garment–body correspondences without relying on mask supervision, demonstrating improved generalization across diverse poses and body shapes.
Despite these advances, existing approaches remain fundamentally image-centric. When applied to videos, they exhibit temporal flickering and identity drift due to the absence of mechanisms for maintaining consistency across frames. In this work, we revisit VTON from an image-to-animation perspective and introduce a video-native diffusion framework with pose-conditioned generation, enabling unified and temporally coherent virtual try-on in dynamic settings.

\noindent \textbf{Diffusion-based Human Animation.}
Recent advances in human and portrait animation have been driven by diffusion models that focus on generating coherent motion from a single reference image. Methods such as Animate Anyone and Champ~\cite{hu2024animate, zhu2024champ} extend 2D UNet architectures~\cite{rombach2022high} with temporal attention layers~\cite{guo2023animatediff}, enabling them to leverage the strong prior knowledge learned from large-scale 2D image models while generating temporally consistent animation videos from only a single image input. Contemporary state-of-the-art avatar animation systems~\cite{tu2025stableanimator, wang2025unianimate, yang2024megactorsigma} further employ video diffusion frameworks to achieve robust pose transfer and strong identity preservation, even under complex motion patterns. DisPose~\cite{li2024dispose} incorporates ControlNet-based keypoint conditioning to provide accurate, pose-driven control for single-image animation. 
Existing animation models, however, lack mechanisms for garment transfer and thus are unable to generate video-based virtual try-on results from a human image paired with a separate garment input.
Our pipeline addresses this gap by integrating VTON with pose-conditioned video animation to produce motion-consistent videos with garment fidelity.

\input{figures/02_overview/figure}
\noindent \textbf{Subject-driven Image and Video Generation.}
Subject-driven generation methods condition on identity while composing additional attributes. Diffusion Transformer–based image synthesis models~\cite{li2025visualcloze, she2025mosaic, wu2025less} place strong emphasis on identity preservation and compositional control, and they can be adapted to virtual try-on through localized inpainting. However, similar to image-based VTON systems, these models still rely on a separate animation stage to achieve temporal coherence.
Recently, diffusion-based models emerge that generate videos directly from a subject~\cite{xu2024anchorcrafter, jiang2025vace, liu2025phantom, yuan2025identity}. These models take a subject image and a text prompt as input and control the subject’s actions and background according to the prompt.
VACE~\cite{jiang2025vace} builds upon a video diffusion transformer~\cite{wan2025wan, hacohen2024ltx} and shows that diverse tasks such as video to video editing and reference to video generation can be unified through a single auxiliary module. 
Through this process, VACE performs reference to video generation followed by pose conditioned video to video synthesis, thereby enabling pose controlled virtual try on.
However, when pose, garment, and human image are jointly conditioned through a single auxiliary module, the model often struggles to preserve fine garment details or to synthesize pose motion accurately. Our approach addresses this limitation by separating pose and garment conditioning into independent network modules and training them jointly. This design improves pose motion synthesis and garment detail accuracy while simultaneously enhancing identity preservation.

%% file: figures/02_overview/figure.tex
\begin{figure*}[t]\centering
\includegraphics[width=\linewidth, trim={0 0 0 0},clip]{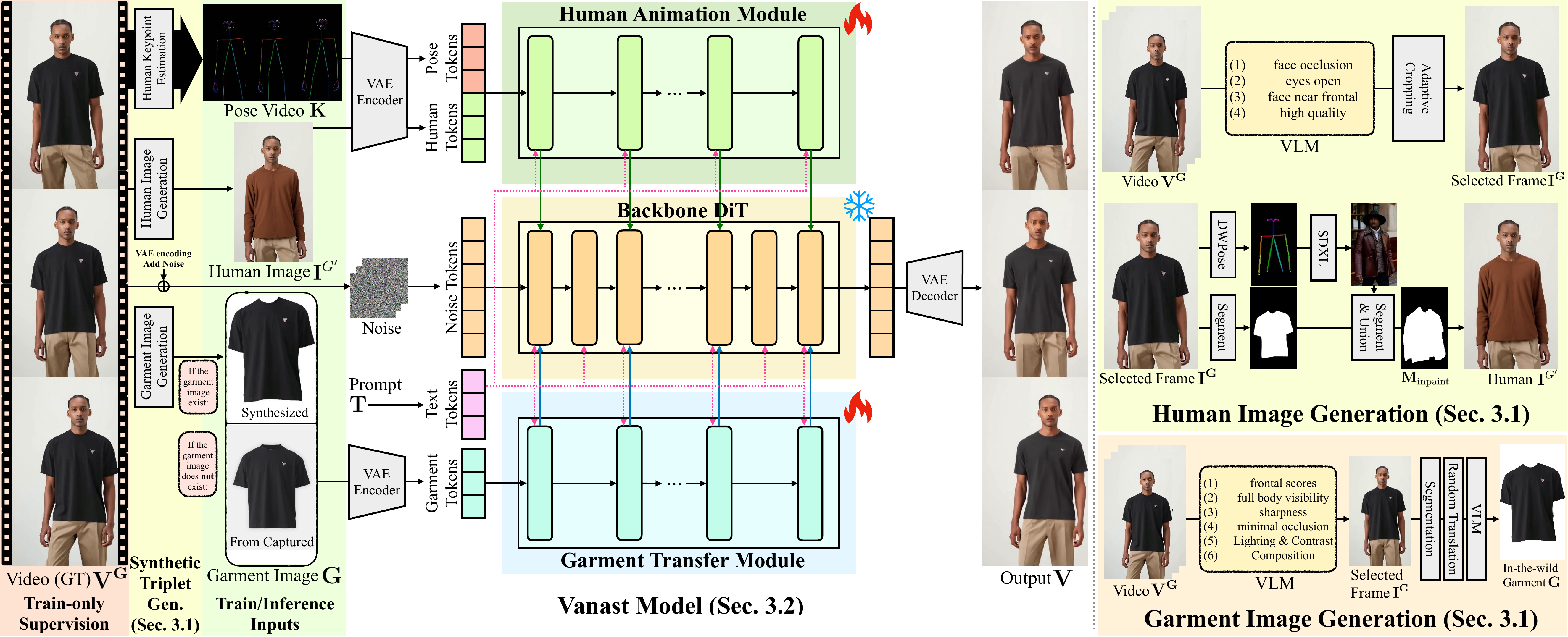}
\vspace{-14px}
\caption{\textbf{Overview of \modelname Pipeline.} 
Our \modelname framework generates virtual try-on human animation videos from a human image, garment images, and a pose video. By incorporating scalable human-image and garment-image generation pipelines, our method avoids dataset-specific constraints and trains effectively at scale. 
The Dual Modules architecture ensures that the three conditioning signals, human image $\mat{I}^{\mat{G'}}$, garment images $\mat{G}$, and pose video $\mat{K}$, are faithfully reflected in the resulting video.
}
\label{fig:overview}
\vspace{-7px}
\end{figure*}

%% file: sections/04_method.tex
\vspace{-1mm}
\section{Method}
\vspace{-1mm}
Our goal is to generate a human animation video with garment transfer in a single stage framework, given a target human image, one or more target garment images, and a motion guidance video, as shown in \cref{fig:teaser}.
Our framework takes, as input, the target garment images $\mat{G}$, a target human image $\mat{I}^{\mat{G'}}$ showing the person wearing an arbitrary garment $\mat{G}' \neq \mat{G}$, a motion guidance video $\mat{K}$, and a text prompt $\mat{T}$ describing human actions and background context. As output, our model  generates an $F$-frame animation sequence denoted as $\mathbf{V} = \{\mat{I}^{\mat{G}}_{t}\}_{t=1}^F$, where each frame $\mat{I}^{\mat{G}}_{t}$ represents a temporally coherent synthesis of the person wearing the target garment items $\mat{G}$. Our model supports multiple garment or accessory items, expressed as $\mat{G} = \{ \mat{G}_1,...,\mat{G}_n\}$, encompassing upper and lower garments and accessories such as hats. We define our model as follows:
\begin{equation}
\mathbf{V} = \mathrm{\modelname}(\mat{G}, \mat{I}^{\mat{G'}}, \mat{K}, \mat{T}).
\end{equation} 
To train \modelname, we construct a triplet dataset consisting of $\mat{I}^{\mat{G'}}$, $\mat{G}$, and a ground truth video $\mathbf{V}^{\mat{G}}$ in which the person is in motion while wearing the target garment $\mat{G}$. The input motion sequence $\mat{K}=\{ \mathbf{k}_t\}_{t=1}^F$ is obtained by applying an off-the-shelf 2D keypoint estimator DWPose~\cite{yang2023effective} to $\mathbf{V}^{\mat{G}}$. Examples from our triplet dataset are shown in \cref{fig:dataset_examples}.

\subsection{Synthetic Triplet Dataset Generation}
Constructing the required triplet dataset ($\mat{I}^{\mat{G'}}$, $\mat{G}$, $\mathbf{V}^{\mat{G}}$) is challenging, and no publicly available dataset provides the necessary quality or structure for our task.
As a key contribution, we introduce a pipeline to generate large quantities of high-quality triplet datasets from multiple data sources.
We begin by collecting garment-video pairs ($\mat{G}$, $\mathbf{V}^{\mat{G}}$) from online shopping platforms, where videos feature diverse identities and backgrounds. 
However, constructing $\mat{I}^{\mat{G'}}$, the target person image wearing an arbitrary garment different from $\mat{G'}$, is non-trivial, because such platforms rarely include images of the same person wearing multiple outfits.
A naive strategy would be to select a frame from $\mathbf{V}^{\mat{G}}$, resulting in $\mat{I}^{\mat{G'}}$ with the identical garment to $\mat{G}$. As shown in our ablation study, this causes the model to overfit to animating the human appearance following motion cues $\mat{K}$, rather than learning garment transfer. 

To address this, we present a synthesis pipeline that generates $\mat{I}^{\mat{G'}}$ using pre-trained image diffusion models, producing realistic images of the same person wearing different garments.
In addition, we introduce a complementary strategy to consgtructs the entire triplet directly from in-the-wild videos $\mathbf{V}^{\mat{G}}$, by automatically extracting both $\mat{G}$ and $\mat{I}^{\mat{G'}}$. This enables us to incorporate more diverse garment appearances that go beyond the clean, studio-style images typically found in online shopping environments.
Finally, we collect our own studio-captured, high-quality dataset to support scenarios involving multiple garments, where $\mat{G} = \{ \mat{G}_1,...,\mat{G}_n\}$. 
Examples of our newly collected multi-garment dataset is shown in \cref{fig:dataset_examples}.

\noindent \textbf{Synthesizing $\mat{I}^{\mat{G'}}$ from ($\mat{G}$, $\mathbf{V}^{\mat{G}}$).}
We leverage a pre-trained diffusion inpainting model, FLUX~\cite{flux2024} to modify only the garment or object regions of $\mat{I}^{\mat{G}}$, where $\mat{I}^{\mat{G}}$ is a selected frame from  $\mathbf{V}^{\mat{G}}$. For high quality synthesis, our pipeline consists of three stages: (1) selecting suitable candidate frames $\mat{I}^{\mat{G}}$, (2) constructing proper and effective inpainting mask $\mat{M}_\mathrm{inpaint}$ to ensure realistic and plausible garment synthesis, and (3) diversifying the synthesized garment type in $\mat{I}^{\mat{G}}$ using LLM-based prompting.

\input{figures/03_dataset_examples/figure}
First, we describe how to select candidate images $\mat{I}^{\mat{G}}$ from $\mathbf{V}^{\mat{G}}$. We randomly sample $n$ frames from each video and select a representative frame using the Vision-Language Model (VLM), Qwen2.5-VL~\cite{bai2025qwen2}. The selected frame satisfies the following criteria: (1) the face is not occluded by hands or objects (sunglasses or masks are allowed); (2) both eyes are open; (3) the face is near-frontal; and (4) the quality score reflecting focus, noise, and exposure is at least $95/100$. If no frame meets these conditions, the first frame of the video is selected. After frame selection, we perform adaptive cropping to support a wide range of human image scales. 
After frame selection, the face and body detection model~\cite{redmon2016you} detects the largest face and full-body bounding boxes.
Each bounding box is expanded according to a predefined scale, and an interpolated bounding box is generated through random linear interpolation between them. A rectangular crop region with a $9:16$ aspect ratio that fully contains the interpolated box is then computed and adjusted to fit within the image boundaries. %

For each selected candidate image $\mat{I}^{\mat{G}}$, we construct a target mask region $\mat{M}_\mathrm{inpaint}$ for the inpainting. A key requirement is that the mask must not simply follow the silhouette of the original garment $\mat{G}$, as the inpainting model tends to preserve the existing garment shape instead of generating a diverse garment for $\mat{I}^{\mat{G'}}$. Thus, the mask should reflect the expected garment region rather than the observed one. Inspired by PERSE~\cite{Cha_2025_CVPR}, we leverage a text-to-image model~\cite{podell2023sdxl} to first synthesize an auxiliary images that maintains the same pose as $\mat{I}^{\mat{G}}$ but features an arbitrary garment and identity. Then, we extract a garment mask from this synthesized image using an off-the-shelf segmentation model~\cite{xie2021segformer}.

Finally, using the chosen $\mat{I}^{\mat{G}}$ and constructed mask $\mat{M}_{\mathrm{inpaint}}$, we generate $\mat{I}^{\mat{G'}}$ as an image of the same person wearing a different garment from the same category as $\mat{G}$.
The text prompts for inpainting are randomly composed from a predefined pool of garment types and colors using ChatGPT~\cite{openai2025chatgpt}. To ensure gender-consistent garment descriptions, we employ the VLM~\cite{bai2025qwen2} to classify each image as male or female and incorporate the result into the text prompt.
Finally, diffusion inpainting model~\cite{flux2024} modifies only the garment or object regions based on the generated masks $\mat{M}_{\mathrm{inpaint}}$ and prompt, producing high-quality human fashion images $\mat{I}^{\mat{G'}}$. 
We show an overview of the human image generation process in \cref{fig:overview}.

\noindent \textbf{Synthesizing $(\mat{I}^{\mat{G'}}, \mat{G})$ from $\mathbf{V}^{\mat{G}}$.} 
To further increase pose and background diversity, we present a pipeline to collect the desired triplet data from an in-the-wild video dataset~\cite{wang2024humanvid}. For constructing $\mat{I}^{\mat{G'}}$, we use the same aforementioned pipeline. However, because in such videos the corresponding garment image $\mat{G}$ is not available, we therefore introduce a method to generate $\mat{G}$ directly from $\mathbf{V}^{\mat{G}}$, enabling the construction of a complete triplet dataset.
We design a process select sutable candidate frame to synthesize garment images $\mat{G}$ from in-the-wild videos $\mathbf{V}^{\mat{G}}$. From each video, $n$ frames are randomly sampled, and frontal scores are obtained using the VLM~\cite{bai2025qwen2}. Among the top $k$ frames with the highest frontal scores, the best frame is selected according to the following priority criteria evaluated by the VLM: (1) full-body visibility (from head to toe), (2) sharpness, (3) minimal occlusion by hands, arms, or objects, (4) lighting and contrast, and (5) composition. The most suitable frame is then chosen. From the selected frame, we extract an upper-clothing mask using the segmentation model~\cite{xie2021segformer}. A garment-highlighted image is generated by filling the background outside the garment mask with white. To prevent the model from being biased by garment position, random translation is applied based on the mask bounding box. Finally, the VLM determines whether each segmented garment or object is valid as the target garment, filtering out unstable segmentations from the segmentation model. This synthetic garment image generation process enables the creation of synthetic triplet datasets from in-the-wild videos, contributing to dataset scale-up and enhancing model robustness. We describe the garment image generation process in \cref{fig:overview}.

\subsection{Model Architecture}
\paragraph{Dual Modules.}
We introduce a training strategy for our video diffusion model using the constructed synthetic triplet dataset. A common approach is to tokenize all conditions and either concatenate or fuse them, but this often leads to slow convergence during training~\cite{jiang2025vace}. Moreover, fine-tuning with a single context module makes it difficult to balance the control of the three conditions~\cite{jiang2023res, jiang2025vace}. 

To address these limitations, we propose a \textbf{Dual Module} architecture. We adopt a distributed and cascaded structure from the backbone T2V (text-to-video) DiT model~\cite{wan2025wan} and divide it into two specialized modules. Inspired by VACE~\cite{jiang2025vace}, we design a Human Animation Module (HAM) that focuses on generating human animation using human and pose images, and a Garment Transfer Module (GTM) that handles garment transfer using garment images. 
HAM and GTM share part of the block architecture with the backbone DiT.
This distributed and cascaded design allows the model to progressively integrate contextual information across multiple levels of representation space, leading to richer conditioning effects compared to single-point injection. The overall formulation is defined as follows:
\begin{equation}
h_{l+1}
=
\begin{cases}
\mathrm{B}^{\text{T2V}}_{l}\!\left( h_{l} \right),
& \text{if } l \neq 2k, \\[6pt]
\begin{aligned}
\mathrm{B}^{\text{T2V}}_{l}\!\left( h_{l} \right)
& {}+ \alpha \cdot \mathrm{B}^{\text{HAM}}_{l}\!\left( h_{l} \right) \\
& {}+ \beta \cdot \mathrm{B}^{\text{GTM}}_{l}\!\left( h_{l} \right),
\end{aligned}
& \text{if } l = 2k.
\end{cases}
\end{equation}
where $B$ denotes each transformer block in the DiT backbone, $l$ is the index of a transformer block, $k$ is a non-negative integer ranging from $0$ to $14$, and $h$ denotes the hidden state which is the input or output of a block. The scalar $\alpha=0.5,\beta=0.5$ controls the relative strength between the HAM and GTM, determining the balance of their contributions during feature integration. We freeze the backbone DiT during training and optimize only the HAM and GTM modules. An overview of our model architecture is presented in \cref{fig:overview}. For detailed model architecture and implementation, refer to supp.mat. Sec.~\textcolor{red}{B}.

\noindent \textbf{Tokenization.}
To provide tokenized inputs for the Dual Modules, we convert each component of the synthetic triplet dataset into latent representations using the pretrained VAE encoder $\mathcal{E}_{\text{VAE}}$~\cite{wan2025wan}. Let $z_{\mathrm{H}}$, $z_{\mathrm{G}}$, and $z_{\mathrm{P}}$ denote the encoded latents of $\mat{I}^{\mat{G'}}, \mat{G}$, and $\mat{K}$, respectively. For the HAM module, we construct a motion-conditioned appearance context by performing frame-wise concatenation of $z_{\mathrm{H}}$ and $z_{\mathrm{P}}$ along the temporal dimension, following previous approach~\cite{jiang2025vace}. For the GTM module, we use $z_{\mathrm{G}}$ alone as input; to match its temporal dimension with HAM, a zero tensor is appended as a placeholder before concatenation. Finally, a 3D convolutional projection layer maps each concatenated latent volume into token embeddings suitable for downstream processing.

\noindent \textbf{Garment interpolation.}
Our model is capable of transferring interpolated garments, constructed from two garments $\mat{G_A}, \mat{G_B}$ belonging to the same category, to the human animation in a zero-shot manner without any additional optimization~\cite{Cha_2025_CVPR, zhang2024diffmorpher}. To achieve this, we obtain the outputs of the GTM transformer blocks for each garment image and compute an $\gamma$-weighted summation of the two representations as follows:
\begin{equation}
\begin{aligned}
h_{l+1}
&=
\mathrm{B}^{\text{T2V}}_{l}\!\left( h_{l} \right)
+
\alpha \cdot \mathrm{B}^{\text{HAM}}_{l}\!\left( h_{l} \right) \\
&\quad
+
\gamma \cdot \mathrm{B}^{\text{GTM}}_{l}\!\left( h_{l};\mat{G_A}\right)
+
(1-\gamma) \cdot \mathrm{B}^{\text{GTM}}_{l}\!\left( h_{l};\mat{G_B} \right),
\end{aligned}
\end{equation}
where $\gamma \in [0, 1]$ denotes the interpolation ratio. This allows the model to produce smooth and semantically coherent interpolations between garments.

%% file: figures/03_dataset_examples/figure.tex
\begin{figure}[t]\centering
\includegraphics[width=\linewidth, trim={0 0 0 0},clip]{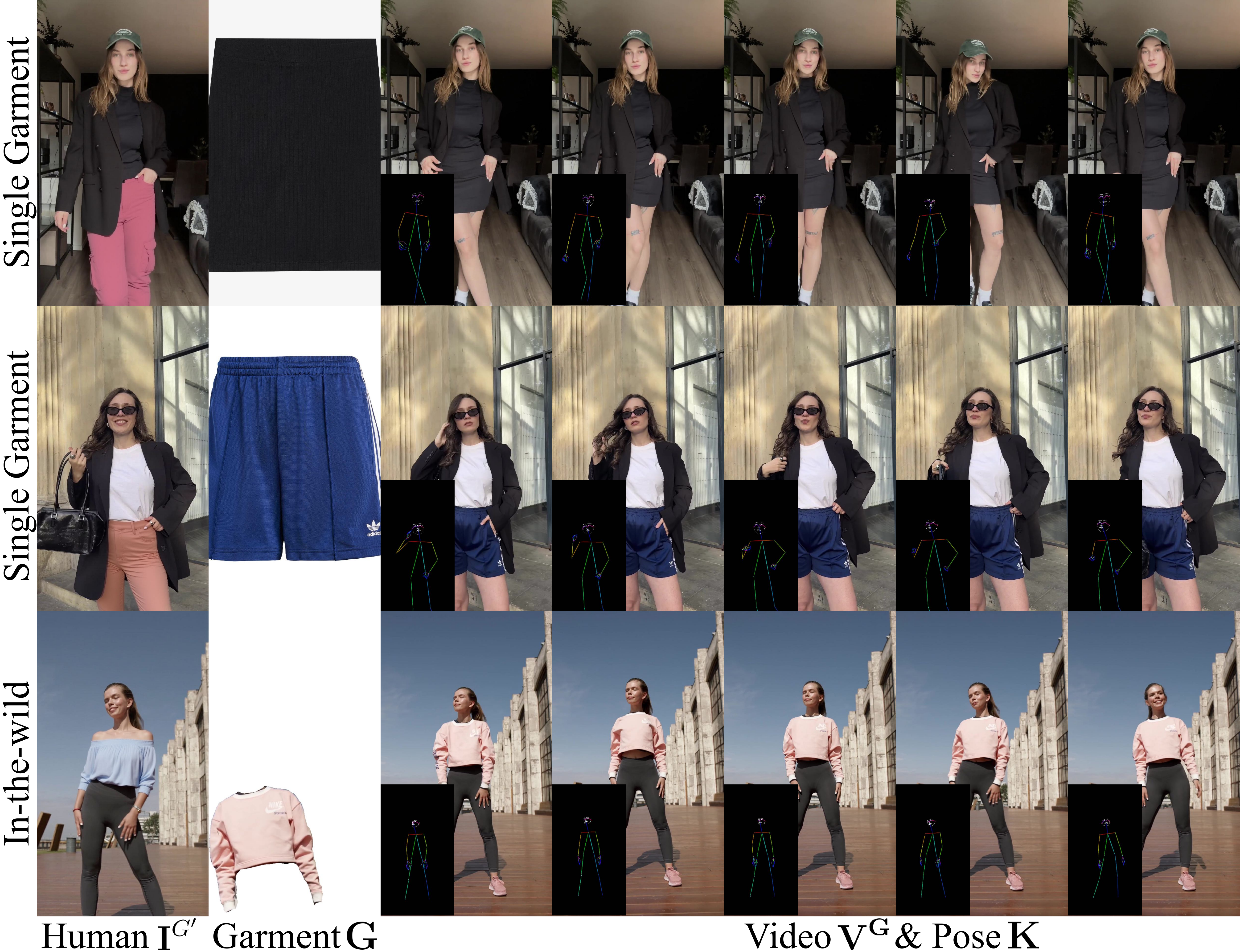}
\vspace{-13px}
\caption{\textbf{Samples of Synthetic Triplet Datasets.} 
We show samples of the datasets used for generation and training. The triplet construction contributes to enabling the model to preserve identity while accurately transferring garments and producing animation videos that follow the target pose.
}
\label{fig:dataset_examples}
\vspace{-10px}
\end{figure}

%% file: sections/05_experiments.tex
\section{Experiments}
\noindent\textbf{Datasets.}
We train our model on a total of $9,135$ videos, each ranging from 3 to 10 seconds in length. These videos are sourced from public internet shopping-mall sites, our captured dataset, and an in-the-wild video dataset~\cite{wang2024humanvid}. 
We construct two evaluation datasets to compare our model with the baselines. The first dataset, referred to as the Internet dataset, is sourced from publicly available shopping-mall websites and consists of videos and standalone garment product images. The second dataset is built using the official test split of the ViViD dataset~\cite{fang2024vivid}. Since the ViViD videos do not contain visible faces, we generate $\mat{I}^{\mat{G'}}$ using an image outpainting model~\cite{flux2024} before leveraging our synthetic triplet dataset generation pipeline.
Although the alignment requirements for human images differ across VTON models, they generally support full-body inputs. Therefore, for evaluation, we use only full-body or near–full-body images as $\mat{I}^{\mat{G'}}$. In total, we randomly sample 80 samples from the Internet dataset and 50 samples from the ViViD dataset. Note that the garment images, human images, and pose videos used for evaluation are entirely disjoint from those in the training set. Further details are provided in the supp.mat.

\noindent\textbf{Metrics.}
We evaluate each method with standard metrics for fidelity and perceptual quality. In particular, we report six metrics: L$_1$, PSNR, SSIM, LPIPS, FID~\cite{heusel2017fid}, and VFID~\cite{fang2024vivid}. 
We apply L1 distance, PSNR, SSIM, and LPIPS frame-wise between generated video and ground-truth. Together, these metrics quantify both the accuracy of the garment transfer and the degree of identity preservation.
We also leverage the Fréchet Inception Distance (FID) and VFID to compute perceptual realism and temporal consistency.

\subsection{Comparisons}
\noindent\textbf{Baselines.}
Since no current method directly produces a virtual try-on video from a human image, a garment image, and a keypoint video, we construct a two-stage pipeline for quantitative comparison. Stage 1 synthesizes a single image of the person wearing the target garment, and Stage 2 animates the image generated in stage 1 using the target motion. We again divide stage 1 into two types, image virtual try-on generation and subject-to-image generation, resulting in 16 model combinations in total. 

The image virtual try-on models we use for garment transfer in Stage 1 are as follows: OOTDiffusion~\cite{xu2025ootdiffusion}, a mask-conditioned dual-Unet diffusion model for reference-guided garment transfer; CatVTON~\cite{chong2024catvton}, a diffusion-based try-on model that supports mask-conditioned synthesis; OmniTry~\cite{feng2025omnitry} and Any2AnyTryon~\cite{guo2025any2anytryon}, diffusion transformer models that operate without explicit masks for reference-guided try-on. Regarding CatVTON, the model produces cropped outputs. 
To match the aspect ratio and resolution across methods, we composite CatVTON results back onto the original image. This may favor pixel-wise metrics by reducing framing discrepancies. 
For subject-to-image generation models we use: VisualCloze~\cite{li2025visualcloze}, MOSAIC~\cite{she2025mosaic}, and UNO~\cite{wu2025less}, diffusion-transformer-based models for subject-conditioned image synthesis including virtual try-on. For human image animation in Stage 2, we employ two models: StableAnimator~\cite{tu2025stableanimator}, video diffusion-based human image animation models; DisPose~\cite{li2024dispose}, a video diffusion human animation model with ControlNet module.

\input{figures/04_qual_s2i/figure}
\input{figures/04_qual_vton/figure}
We additionally evaluate VACE~\cite{jiang2025vace}, a diffusion transformer-based model for subject-driven video generation. While VACE does not explicitly support a single-stage try-on pipeline, its unified architecture can accept multiple conditioning signals simultaneously. Therefore, we also evaluate VACE in a single-stage inference setting, where the model receives a pose video, a human image, and a garment image as joint inputs to directly produce the final video. For completeness, we additionally assess a two-step variant in which we first generate a video of the person wearing the garment from the human and garment images, and then feed the first frame of that video together with the pose video to obtain the final animation.

\input{tables/01_quant_subject_gen}

\input{tables/03_quant_vton}

\noindent\textbf{Results.}
As shown in \cref{tab:quant_subject_gen}, our model achieves the best performance across all metrics when compared with combinations of subject-to-image generation models and animation models. Qualitative results in \cref{fig:qual_s2i} further confirm that our approach produces the most accurate pose following and garment transfer, while preserving identity more faithfully than all subject-to-image–based baselines.

As presented in \cref{tab:quant_vton}, our model also outperforms all combinations of virtual try-on and animation models on every metric except SSIM, for which our score remains comparable to the best-performing method. Qualitative comparisons in \cref{fig:qual_vton} demonstrate that our results most closely resemble the ground truth among all image virtual try-on–based baselines. Additional results are provided in the supplemental materials.
\input{figures/06_single_garment/figure}

\subsection{Ablation Study}
We evaluate the contributions of our key architectural components as well as the effectiveness of the synthetic triplet dataset. Quantitative results are reported in \cref{tab:ablation_study}, and qualitative comparisons are shown in \cref{fig:ablation}. \textbf{``Single Module''} refers to a baseline in which the T2V backbone is frozen and a single trainable module is used. All conditions from the triplet dataset are concatenated and fed into this module for training. \textbf{``Backbone-LoRA''} denotes a model that directly concatenates all input conditions into the T2V backbone without introducing any additional modules, and fine-tunes the model using LoRA layers, applied to every DiT block. This setup enables faster convergence while preserving the generative capabilities of T2V. \textbf{``w/o SynthHuman''} shares the same architecture as our full model but is trained without using $I^{G'}$, relying solely on $I^{G}$.

As shown in \cref{tab:ablation_study}, our model achieves the highest performance across all metrics. The qualitative results in \cref{fig:ablation} further demonstrate that our outputs most closely resemble the ground truth. The ``Single Module'' baseline fails to accurately control pose conditions, while the ``Backbone-LoRA'' and ``w/o SynthHuman'' variants struggle to perform correct garment transfer.
\input{tables/02_ablation_study}
\input{figures/05_ablation/figure}

\subsection{Application}
\noindent\textbf{Garment interpolation.}
As shown in \cref{fig:interpolation}, our model is capable of generating human animation videos in a single stage without any additional training, while interpolating and transferring garments between two input garments according to the interpolation weight $\alpha$. Both upper and lower garments exhibit smooth and natural interpolation.

\noindent\textbf{Multiple garment transfer.}
Our model supports single-garment transfer, as illustrated in \cref{fig:single_garment}, and further enables multiple-garment transfer through the GTM trained on the synthetic triplet dataset.
As shown in \cref{fig:multi_garment}, our approach can transfer both upper and lower garments simultaneously and generate human image animations without any additional training. The results demonstrate that garment details for both regions are well preserved while maintaining strong identity consistency.

\noindent\textbf{In-the-wild garment transfer.}
Benefiting from the combination of the in-the-wild dataset and the synthetic triplet dataset, our model is able to perform garment transfer from in-the-wild garment images. As shown in \cref{fig:in_the_wild}, the model successfully transfers garments despite the mismatch between the garment image pose and the human pose during animation, while maintaining strong temporal consistency throughout the generated sequence.
\input{figures/07_multi_garment/figure}
\input{figures/08_in_the_wild/figure}
\input{figures/09_interpolation/figure}

%% file: figures/04_qual_s2i/figure.tex
\begin{figure}[t]\centering
\includegraphics[width=\linewidth, trim={0 0 0 0},clip]{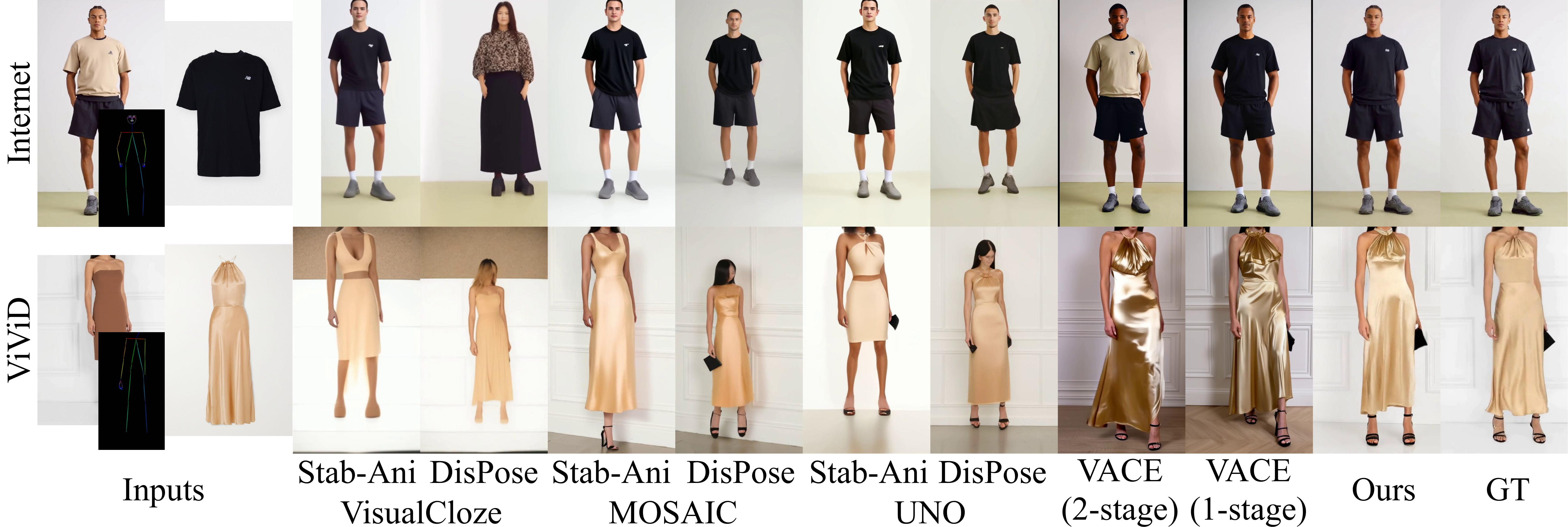}
\vspace{-13px}
\captionof{figure}{\textbf{Qualitative Comparisons (Subject-to-Image-based).} 
We compare our results with baselines constructed by combining subject-to-image generation and animation models. Our method produces the most accurate pose following and garment transfer while preserving identity with high fidelity.
}
\label{fig:qual_s2i}
\vspace{-10px}
\end{figure}

%% file: figures/04_qual_vton/figure.tex
\begin{figure}[t]\centering
\includegraphics[width=\linewidth, trim={0 0 0 0},clip]{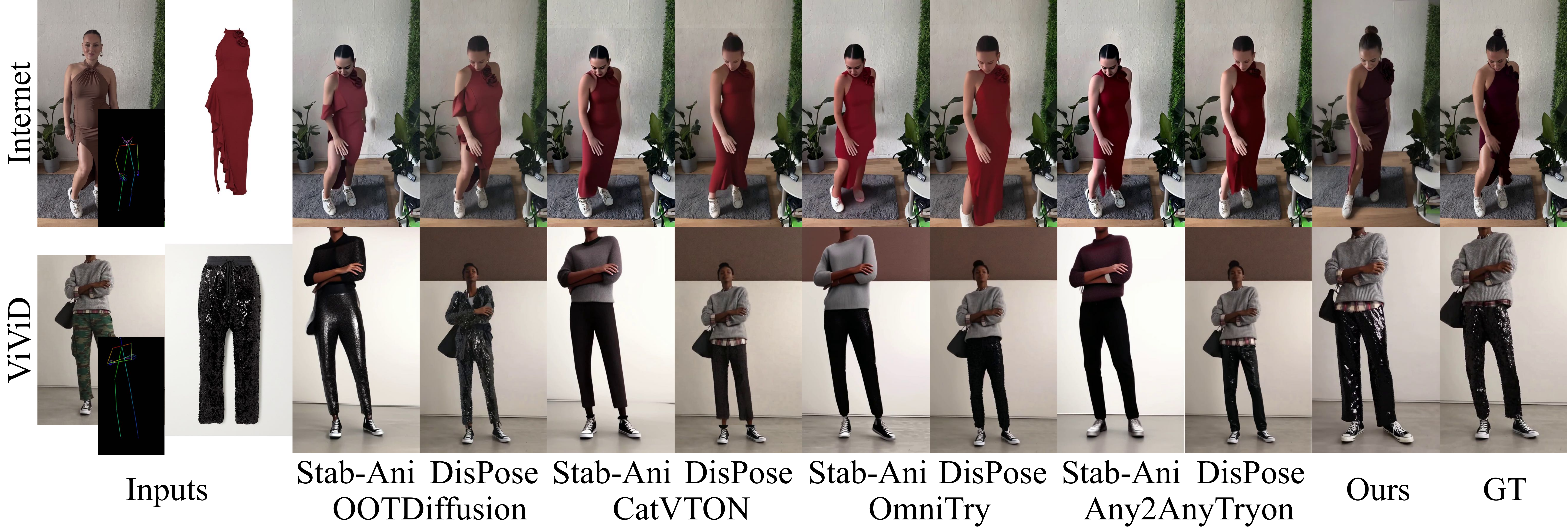}
\vspace{-13px}
\captionof{figure}{\textbf{Qualitative Comparisons (Virtual Try-On-based).} 
We compare our results with baselines formed by combining image virtual try-on models with animation models. Our method achieves the most accurate pose following and garment transfer while preserving identity with the highest fidelity.
}
\label{fig:qual_vton}
\vspace{-10px}
\end{figure}

%% file: tables/01_quant_subject_gen.tex
\begin{table*}[t]
  \centering
  \vspace{-3mm}
  \caption{\textbf{Quantitative Comparison with the Combination of Subject-to-Image and Animation Models.} 
  We compare our model with a baseline that combines a subject-to-image model and an animation model. Our model achieves the best performance across all metrics. \textbf{Bold text} indicates the best score in each column.
  }
  \vspace{-3mm}
  \label{tab:quant_subject_gen}
  \resizebox{\linewidth}{!}{ 
  \begin{tabular}{ll|ccccccc|ccccccc}
    \toprule
    \multicolumn{2}{l|}{} & \multicolumn{7}{c|}{\textbf{Internet Dataset}} & \multicolumn{7}{c}{\textbf{ViViD Dataset}} \\
    \cmidrule(lr){3-9} \cmidrule(lr){10-16}
    Img.Gen. & Animation & 
    L$_1\downarrow$ & PSNR$\uparrow$ & SSIM$\uparrow$ & LPIPS$\downarrow$ & FID$\downarrow$ & VFID$_\text{I3D}\downarrow$ & VFID$_\text{ResNeXt}\downarrow$ & 
    L$_1\downarrow$ & PSNR$\uparrow$ & SSIM$\uparrow$ & LPIPS$\downarrow$ & FID$\downarrow$  & VFID$_\text{I3D}\downarrow$ & VFID$_\text{ResNeXt}\downarrow$ \\
    \midrule
    \multirow{2}{*}{VisualCloze}
      & StableAnimator      
      & 0.2266 & 11.08 & 0.7210 & 0.4676 & 210.37 & 34.48 & 1.69 
      & 0.2835 & 8.80 & 0.6355 & 0.6061 & 190.83 & 49.04 & 7.21 \\
      & DisPose             
      & 0.2590 & 9.49 & 0.6527 & 0.5884 & 205.68 & 40.49 & 1.93 
      & 0.3043 & 8.25 & 0.5880 & 0.6871 & 214.00 & 55.82 & 11.71 \\
    \midrule
    \multirow{2}{*}{MOSAIC}
      & StableAnimator      
      & 0.1875 & 12.20 & 0.7128 & 0.4400 & 158.76 & 32.36 & 2.34 
      & 0.2382 & 10.02 & 0.6617 & 0.5583 & 158.47 & 44.46 & 2.55 \\
      & DisPose             
      & 0.1714 & 12.36 & 0.7133 & 0.4641 & 155.18 & 35.24 & 1.54
      & 0.2008 & 10.67 & 0.6686 & 0.5619 & 155.98 & 50.90 & 4.91\\
    \midrule
    \multirow{2}{*}{UNO}
      & StableAnimator      
      & 0.2025 & 11.79 & 0.7148 & 0.4381 & 162.15 & 31.73 & 3.12 
      & 0.2556 & 9.63 & 0.6622 & 0.5577 & 158.31 & 45.39 & 2.95 \\
      & DisPose             
      & 0.1774 & 12.06 & 0.7071 & 0.4734 & 154.27 & 34.08 & 1.65
      & 0.2125 & 10.41 & 0.6610 & 0.5609 & 140.26 & 45.55 & 3.03 \\
    \midrule
    \multicolumn{2}{l|}{VACE (2-stage)} 
    & 0.1708 & 12.44 & 0.6618 & 0.4507 & 141.89 & 49.87 & 5.44
    & 0.1994 & 11.03 & 0.6053 & 0.5678 & 143.00 & 46.22 & 4.59 \\
    \multicolumn{2}{l|}{VACE (1-stage)} 
    & 0.1453 & 13.09 & 0.6894 & 0.4052 & 115.40 & 47.31 & 5.86 
    & 0.1733 & 11.62 & 0.6245 & 0.5363 & 134.96 & 43.97 & 3.20 \\
    \midrule
    \multicolumn{2}{l|}{\textbf{Ours}} 
    & \textbf{0.0719} & \textbf{17.95} & \textbf{0.7550} & \textbf{0.2370} & \textbf{91.05} & \textbf{22.52} & \textbf{0.39}
    & \textbf{0.1077} & \textbf{14.67} & \textbf{0.6686} & \textbf{0.3649} & \textbf{105.89} & \textbf{35.72} & \textbf{1.30} \\
    \bottomrule
  \end{tabular}
  }
\end{table*}

%% file: tables/03_quant_vton.tex
\begin{table*}[t]
  \centering
  \vspace{-3mm}
  \caption{\textbf{Quantitative Comparison with the Combination of Image Virtual Try-On and Animation Models.} 
  We compare our model with a baseline that combines a image virtual try-on model and an animation model. Our model achieves the best performance across all metrics. \textbf{Bold text} indicates the best score in each column.
  }
  \vspace{-3mm}
  \label{tab:quant_vton}
  \resizebox{\linewidth}{!}{ 
  \begin{tabular}{ll|ccccccc|ccccccc}
    \toprule
    \multicolumn{2}{l|}{} & \multicolumn{7}{c|}{\textbf{Internet Dataset}} & \multicolumn{7}{c}{\textbf{ViViD Dataset}} \\
    \cmidrule(lr){3-9} \cmidrule(lr){10-16}
    Img.Gen. & Animation & 
    L$_1\downarrow$ & PSNR$\uparrow$ & SSIM$\uparrow$ & LPIPS$\downarrow$ & FID$\downarrow$ & VFID$_\text{I3D}\downarrow$ & VFID$_\text{ResNeXt}\downarrow$ & 
    L$_1\downarrow$ & PSNR$\uparrow$ & SSIM$\uparrow$ & LPIPS$\downarrow$ & FID$\downarrow$ & VFID$_\text{I3D}\downarrow$ & VFID$_\text{ResNeXt}\downarrow$ \\
    \midrule
    \multirow{2}{*}{OOTDiffusion} 
      & StableAnimator     
      & 0.1305 & 13.91 & 0.7412 & 0.3428 & 174.79 & 33.17 & 2.55
      & 0.2431 & 9.99 & 0.6566 & 0.5528 & 168.46 & 43.39 & 2.66 \\
      & DisPose            
      & 0.1143 & 14.86 & 0.7551 & 0.3325 & 172.94 & 31.48 & 1.68
      & 0.2101 & 10.72 & 0.6421 & 0.5927 & 169.92 & 45.13 & 8.32 \\
    \midrule
    \multirow{2}{*}{CatVTON}        %
      & StableAnimator     
      & 0.1242 & 14.56 & 0.7649 & 0.3273 & 132.09 & 26.43 & 0.90
      & 0.2415 & 10.04 & \textbf{0.6772} & 0.5441 & 167.27 & 43.12 & 4.62 \\
      & DisPose            
      & 0.0987 & 15.87 & \textbf{0.7779} & 0.3238 & 120.69 & 25.00 & 1.05
      & 0.2083 & 10.66 & 0.6430 & 0.6018 & 155.30 & 43.82 & 9.62 \\
    \midrule
    \multirow{2}{*}{OmniTry}
      & StableAnimator     
      & 0.1227 & 14.53 & 0.7671 & 0.3178 & 121.04 & 25.66 & 1.18
      & 0.2372 & 10.10 & 0.6726 & 0.5425 & 161.33 & 43.36 & 4.21 \\
      & DisPose            
      & 0.0968 & 15.46 & 0.7775 & 0.3151 & 111.08 & 24.60 & 0.97
      & 0.2027 & 10.75 & 0.6496 & 0.5867 & 149.84 & 43.76 & 8.45 \\
    \midrule
    \multirow{2}{*}{Any2AnyTryon}
      & StableAnimator     
      & 0.1250 & 14.37 & 0.7596 & 0.3271 & 126.15 & 26.28 & 1.36
      & 0.2460 & 9.78 & 0.6685 & 0.5518 & 165.04 & 45.13 & 4.26 \\
      & DisPose            
      & 0.0950 & 15.53 & 0.7778 & 0.3160 & 116.39 & 25.39 & 0.78
      & 0.2083 & 10.58 & 0.6426 & 0.5969 & 154.38 & 43.69 & 8.07 \\
    \midrule
    \multicolumn{2}{l|}{\textbf{Ours}} 
    & \textbf{0.0719} & \textbf{17.95} & 0.7550 & \textbf{0.2370} & \textbf{91.05} & \textbf{22.52} & \textbf{0.39}
    & \textbf{0.1077} & \textbf{14.67} & 0.6686 & \textbf{0.3649} & \textbf{105.89} & \textbf{35.72} & \textbf{1.30} \\
    \bottomrule
  \end{tabular}
  }
\end{table*}

%% file: figures/06_single_garment/figure.tex
\begin{figure}[t]\centering
\includegraphics[width=\linewidth, trim={0 0 0 0},clip]{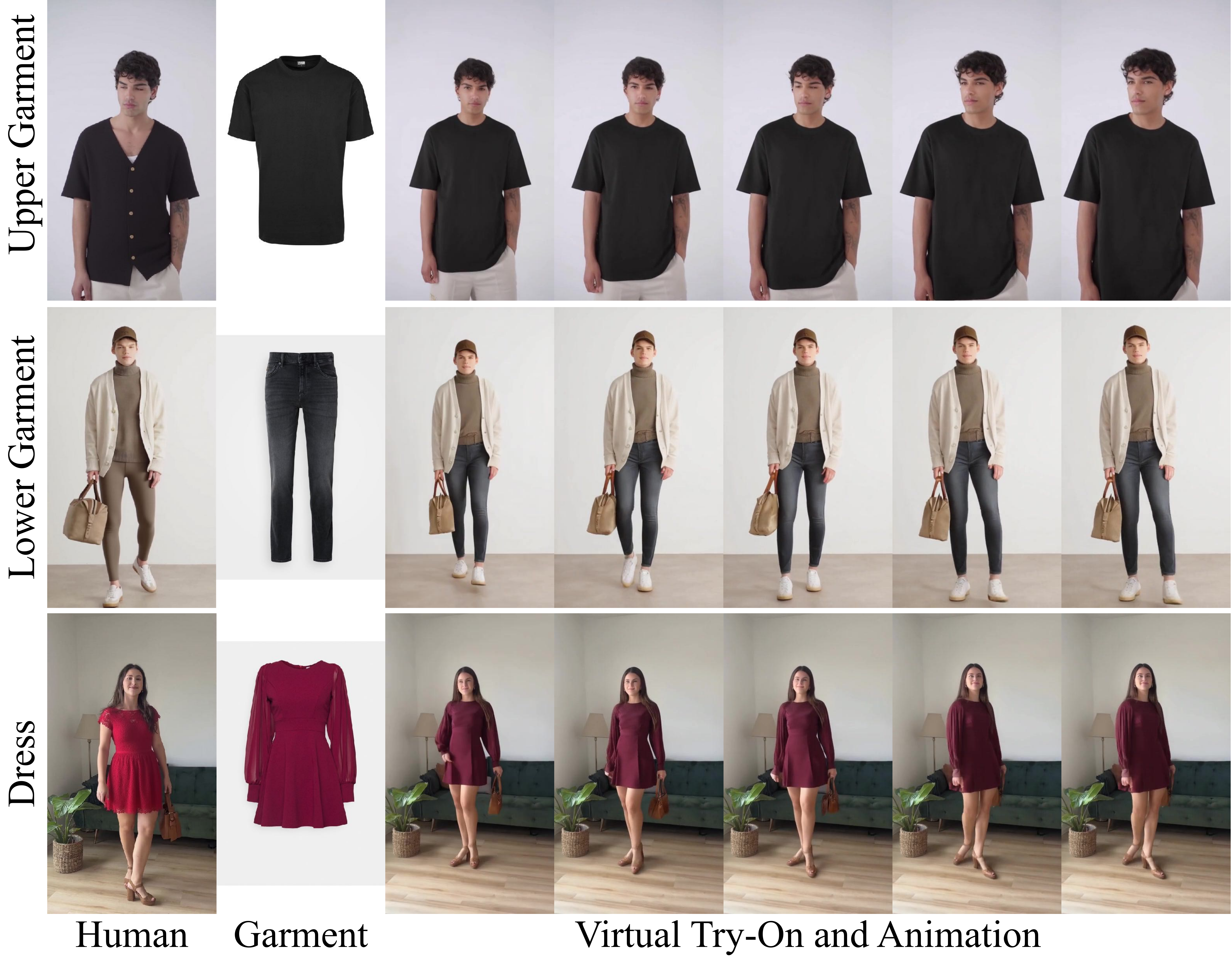}
\vspace{-13px}
\captionof{figure}{\textbf{Result of Single Garment Transfer.} 
We present virtual try-on with human image animation results generated from a single garment image.
}
\label{fig:single_garment}
\vspace{-4px}
\end{figure}

%% file: tables/02_ablation_study.tex
\begin{table}[t]
\centering
\vspace{-3mm}
\caption{\textbf{Ablation Study.}
We conduct ablation study for each component of our model and dataset configuration. \textbf{Bold text} indicates the best score in each column.
}
\vspace{-3mm}
\resizebox{1.0\columnwidth}{!}{
\begin{tabular}{l|cccc|ccc}
\toprule
Method & \thead{L${_1\downarrow}$} & \thead{PSNR${\uparrow}$} & \thead{SSIM${\uparrow}$} & \thead{LPIPS${\downarrow}$} & \thead{FID${\downarrow}$} & 
VFID$_\text{I3D}\downarrow$ & VFID$_\text{ResNeXt}\downarrow$ \\
\midrule
Single Module               
& 0.1162 & 14.28 & 0.6609 & 0.3974 & 108.84 & 39.64 & 1.76 \\
Backbone-LoRA               
& 0.1359 & 13.17 & 0.6314 & 0.4502 & 120.97 & 42.47 & 1.87 \\
w/o SynthHuman              
& 0.1163 & 14.62 & 0.6653 & 0.3943 & 110.76 & 38.89 & 1.93 \\
\midrule
Ours                        
& \bf{0.1069} & \bf{14.74} & \bf{0.6657} & \bf{0.3673} & \bf{104.59} & \bf{35.60} & \bf{1.21} \\
\bottomrule
\end{tabular}
}
\vspace{-5px}
\label{tab:ablation_study}
\end{table}

%% file: figures/05_ablation/figure.tex
\begin{figure}[t]\centering
\includegraphics[width=\linewidth, trim={0 0 0 0},clip]{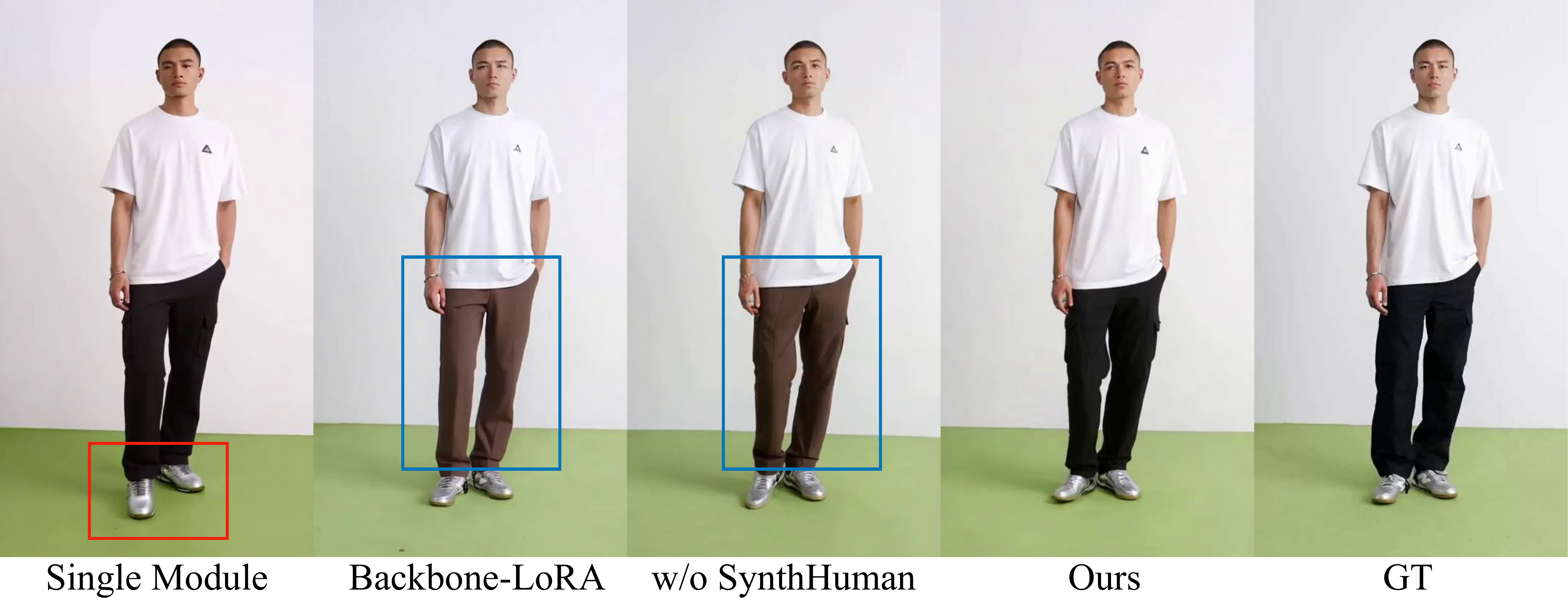}
\vspace{-13px}
\captionof{figure}{\textbf{Ablation Study.} 
We present the ablation study results for the lower garment transfer. The red box in the ``Single Module'' result demonstrates vulnerability to pose conditions. Both ``Backbone-LoRA'' and ``w/o SynthHuman'' fail to achieve accurate garment transfer, as indicated in blue box. In contrast, our full model produces results most similar to the ground truth. 
}
\label{fig:ablation}
\vspace{-4px}
\end{figure}

%% file: figures/07_multi_garment/figure.tex
\begin{figure}[t]\centering
\includegraphics[width=\linewidth, trim={0 0 0 0},clip]{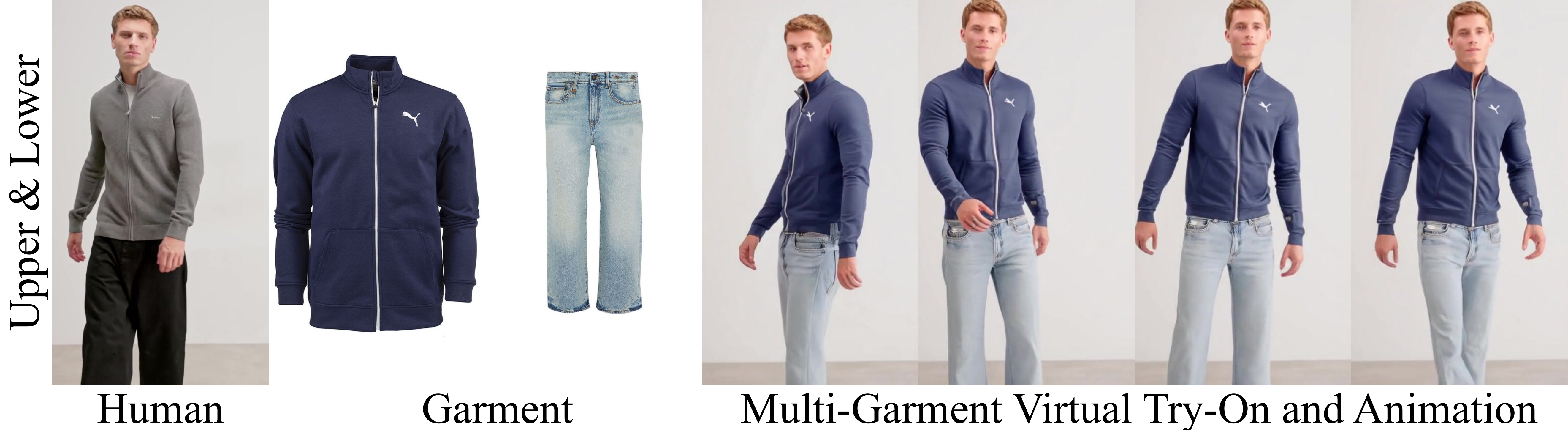}
\vspace{-13px}
\captionof{figure}{\textbf{Result of Multiple Garment Transfer.} 
We present zero-shot garment transfer results where both upper and lower garments are transferred simultaneously. The logos and fine details of the garments are well preserved and accurately reflected in the generated animation videos.
}
\label{fig:multi_garment}
\vspace{-4px}
\end{figure}

%% file: figures/08_in_the_wild/figure.tex
\begin{figure}[t]\centering
\includegraphics[width=\linewidth, trim={0 0 0 0},clip]{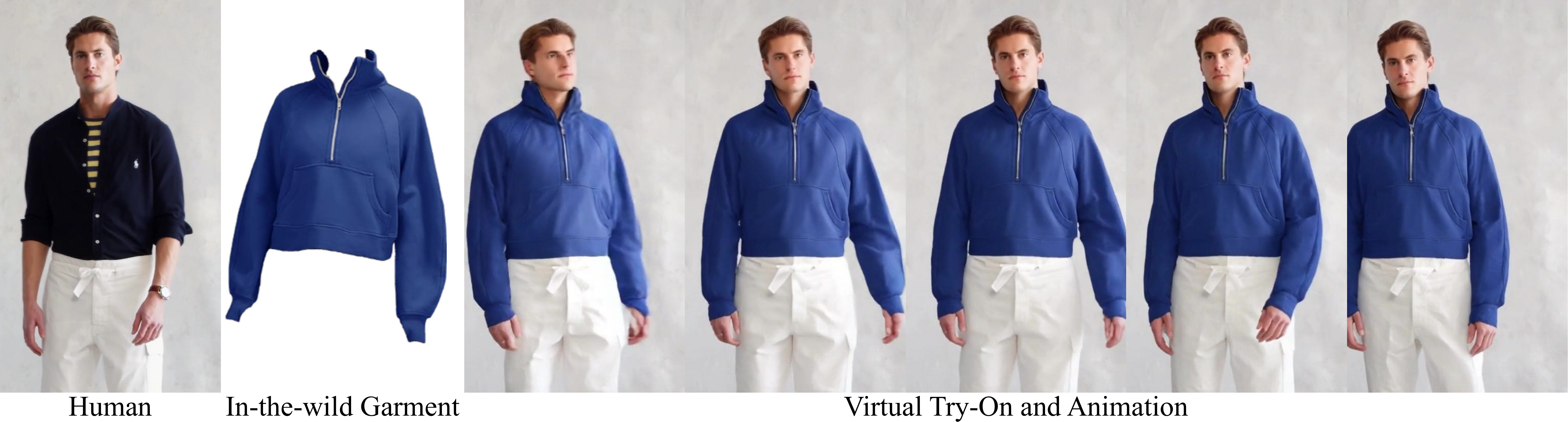}
\vspace{-13px}
\captionof{figure}{\textbf{Result of In-the-wild Garment Transfer.} 
We present garment transfer results using in-the-wild garment images provided by the TikTokDress~\cite{nguyen2025swifttry} dataset.
}
\label{fig:in_the_wild}
\vspace{-5px}
\end{figure}

%% file: figures/09_interpolation/figure.tex
\begin{figure}[t]\centering
\includegraphics[width=\linewidth, trim={0 0 0 0},clip]{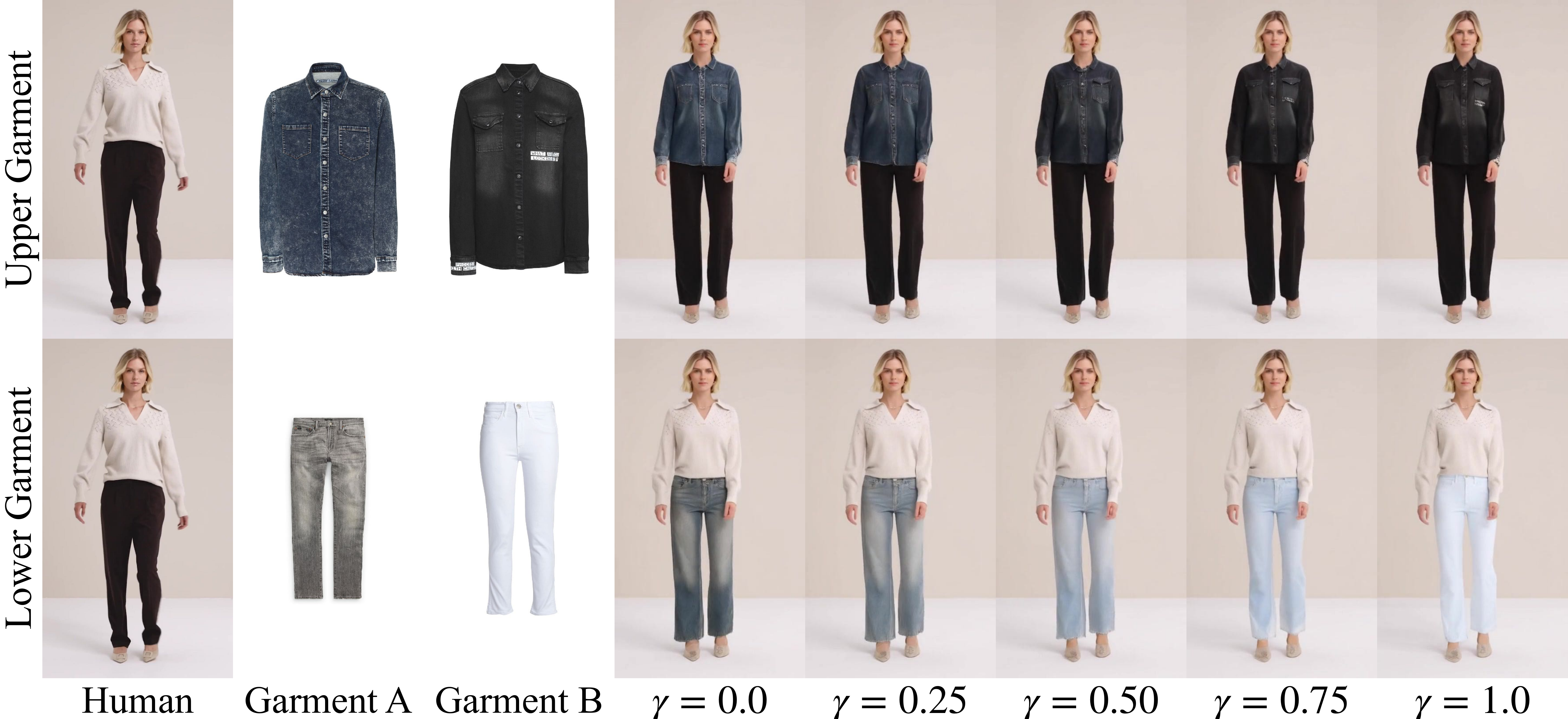}
\vspace{-14px}
\captionof{figure}{\textbf{Result of Garment Interpolation.} 
Without requiring any additional finetuning, our \modelname model performs zero-shot transfer of interpolated garments by GTM. $\gamma$ denotes a scalar interpolation weight.
}
\label{fig:interpolation}
\end{figure}

%% file: sections/06_conclusion.tex
\section{Conclusion}
We introduce \modelname, a unified framework that synthesizes garment-transferred human animation videos directly from a single human image, garment images, and a pose guidance video. 
By constructing scalable triplet supervision from in-the-wild videos and complementary upper–lower garment captures, our pipeline addresses the structural limitations of existing online data and enables identity-preserving training with diverse garment variations.
The proposed Dual Module architecture significantly improves garment fidelity, pose adherence, and identity preservation while supporting zero-shot garment interpolation and multi-garment transfer without finetuning. Extensive experiments show that \modelname consistently surpasses two-stage pipelines combining state-of-the-art virtual try-on and subject-to-image with human animation models, and generalizes effectively to in-the-wild garment images with strong temporal coherence. 

%% file: sections/08_ack.tex
\newpage
\noindent\textbf{Acknowledgments.}
We trained the \modelname model using the Pixophilia-SNU dataset. This work was supported by Samsung Electronics MX Division, NRF grant funded by the Korean government (MSIT) (No. RS-2022-NR070498), and IITP grant funded by the Korea government (MSIT) [No. RS-2024-00439854, No. 2022-0-00156, No. RS-2025-25441838, and No. RS-2021-II211343]. H. Joo is the corresponding author.